\documentclass[conference]{IEEEtran}
\IEEEoverridecommandlockouts
\usepackage{cite}
\usepackage{amsmath,amssymb,amsfonts}
\usepackage{algorithmic}
\usepackage{graphicx}
\usepackage{textcomp}
\usepackage{xcolor}
\def\BibTeX{{\rm B\kern-.05em{\sc i\kern-.025em b}\kern-.08em
    T\kern-.1667em\lower.7ex\hbox{E}\kern-.125emX}}

\usepackage{flushend}
\usepackage{tabularx}
\usepackage[colorlinks=true,linkcolor=black,anchorcolor=black,citecolor=black,filecolor=black,menucolor=black,runcolor=black,urlcolor=black]{hyperref}
\usepackage{svg}

\usepackage{cite}

\DeclareRobustCommand{\IEEEauthorrefmark}[1]{\smash{\textsuperscript{\footnotesize #1}}}

\begin{document}

% guter titel für Saras urlaubsko0nferenz auf der MIE26 (sollte aber eigentlich ein Journal Paper werden)
% \title{Atypical Mytotic Figure Classification in Histopathology with Deep Ensemble Learning and Rule Based Refinement}

\title{Classifying Mitotic Figures in the MIDOG25 Challenge with Deep Ensemble Learning and Rule Based Refinement}

\author{
  \IEEEauthorblockN{
   Sara Krauss\IEEEauthorrefmark{1},
   Ellena Spieß\IEEEauthorrefmark{2,3}, 
   Daniel Hieber\IEEEauthorrefmark{2,3,4},
   Frank Kramer\IEEEauthorrefmark{1},
   Johannes Schobel\IEEEauthorrefmark{3},
   Dominik Müller\IEEEauthorrefmark{1}
  }
\IEEEauthorblockA{\textit{\IEEEauthorrefmark{1} IT-Infrastructure for Translational Medical Research, Faculty of Applied Computer Science, University of Augsburg, Germany}}
\IEEEauthorblockA{\textit{\IEEEauthorrefmark{2} Department of Neuropathology, Pathology, Medical Faculty, University of Augsburg, Germany}}
\IEEEauthorblockA{\textit{\IEEEauthorrefmark{3} DigiHealth Institute, Neu-Ulm University of Applied Sciences, Germany}}
\IEEEauthorblockA{\textit{\IEEEauthorrefmark{4} Institute of Medical Data Science, University Hospital Würzburg, Germany}}
\IEEEauthorblockA{E-Mail: dominik.mueller@informatik.uni-augsburg.de}
}

\maketitle

\begin{abstract}
Mitotic Figures (MFs) are relevant biomarkers in tumor grading. Differentiating Atypical MFs (AMFs) from Normal MFs (NMFs) remains difficult, as manual annotation is time-consuming and subjective. In this work an ensemble of ConvNeXtBase models was trained with AUCMEDI and extend with a rule-based refinement (RBR) module. On the MIDOG25 preliminary test set, the ensemble achieved a balanced accuracy of 84.02\%. While the RBR increased specificity, it reduced sensitivity and overall performance. The results show that deep ensembles perform well for AMF classification. RBR can increase specific metrics but requires further research.
\end{abstract}

\begin{IEEEkeywords}
MIDOG25, Mitosis; Computer Vision, Classification % add mesh: https://www.ncbi.nlm.nih.gov/mesh/
\end{IEEEkeywords}

%##############################################################################
\section{Introduction}
\label{sec:intro}
Mitotic Figures (MFs) have proven their value as a biomarker for tumor proliferation and outcome prediction in some tumors \cite{van_dooijeweert_grading_2022,bertram_mitotic_2024}. Despite AI advances, methods for their analysis and classification remain immature for clinical adoption, and manual annotation remains the standard — a process both time-consuming and vulnerable to inter-observer variance \cite{ibrahim_improving_2023}. The distinguishing of Atypical MF (AMFs) from Normal MF (NMFs) has recently become a new research focus, as AMFs are believed to be another valuable prognostic marker in tumors \cite{lashen_characteristics_2022}. The MIDOG25 Challenge advances research in this field, by providing large datasets of MFs, classified into NMFs and AMFs, serving a valuable foundation for new automated classification methods \cite{ammeling_mitosis_2025}. In this work a new Deep Learning (DL) ensemble is trained on the now available datasets and combined with a rule based refinement (RBR) approach, generating a novel classification pipeline.

%##############################################################################
\section{Methods}
\label{sec:methods}
\begin{figure}[tbp]
    \centerline{\includegraphics[width=0.45\textwidth]{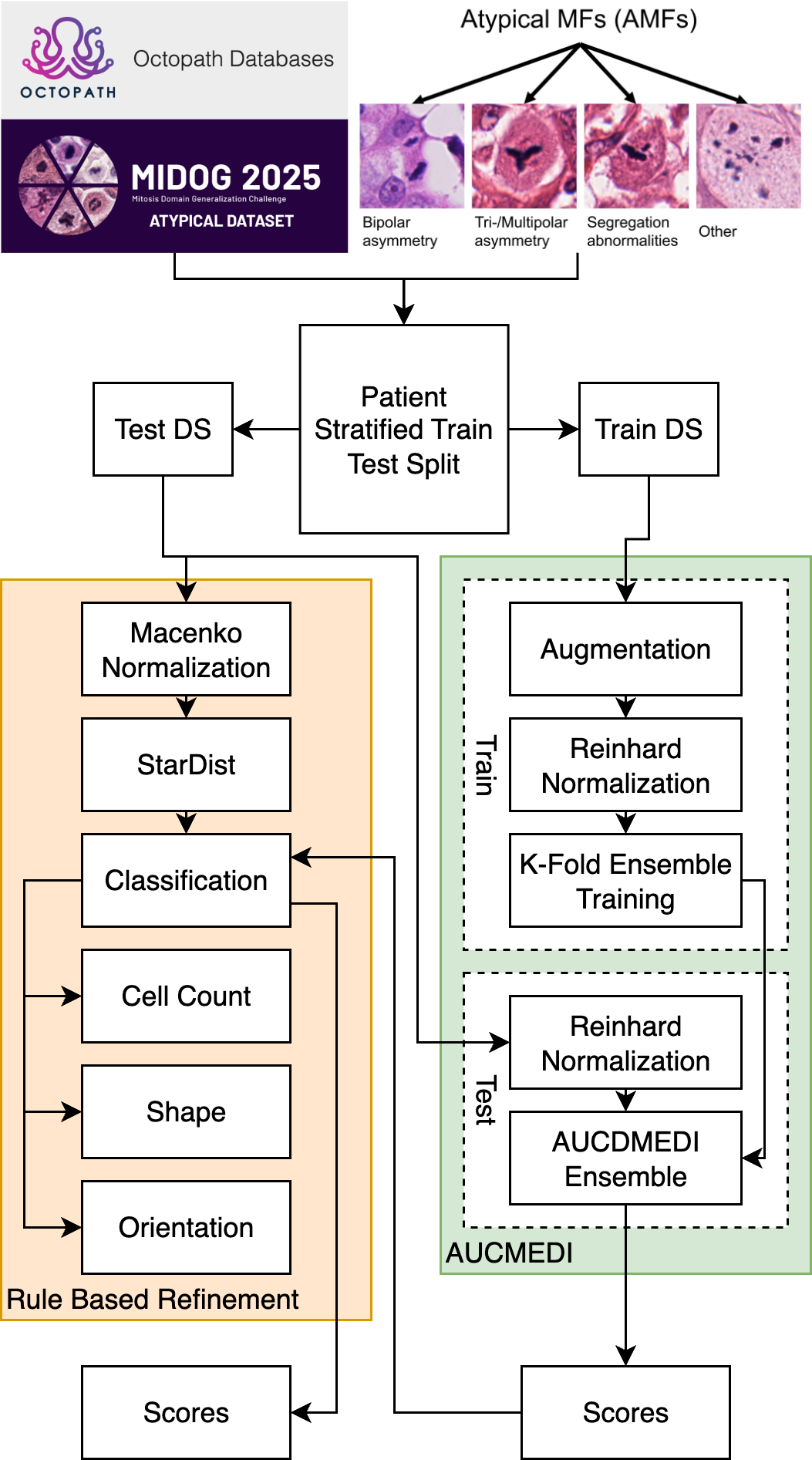}}
    \caption{Training and validation workflow of the proposed classification architecture.}
    \label{fig:pipeline_train}
\end{figure}

The classification process follows a bipartite approach utilizing a DL based classification and RBR as an additional improvement step. Both approaches can be used independently, however only the DL approach is capable of providing classifications for all images, while the RBR can only provide classifications for some images. Figure \ref{fig:pipeline_train} shows an overview of the workflow for training and testing.

All available datasets from the MIDOG25 challenge were used as data. These include the AMi-Br dataset \cite{palm_histologic_2025}, the MIDOG25 dataset \cite{weiss_dataset_2025}, and the Octopath dataset \cite{shen_omg-octo_2025}. All of these datasets contain cropped images of MFs in the center region of Hematoxylin and Eosin (HE)-stained histopathological images. As the AMi-Br and MIDOG25 datasets overlap to some extent, all images already included in the AMi-Br dataset were removed from the MIDOG25 dataset to prevent overfitting. The combined dataset consists of 15,689 MFs from 667 patients. A patient-stratified train-test split of 80/20 was then applied, resulting in 12,431 from 533 patients and 3,258 tiles from 134 patients for training and testing, respectively.

Since the datasets were already pre-processed, additional preprocessing was kept minimal. Only the Reinhard algorithm \cite{reinhard_color_2001} implemented in the torchstain framework \cite{barbano_torchstain_2022}. was used for HE stain normalization. Image \#2935 of the MIDOG25 dataset was set as the target image. An augmentation pipeline was used to increase datasets diversity. These augmentations include the transformations: flip, rotate, brightness, contrast, saturation, hue, scale, gaussian noise, and elastic transformation.

Using the AUCMEDI framework \cite{mayer_standardized_2022} an ensemble of three pre-trained ConvNeXtBase models \cite{liu_convnet_2022} was created. The training was set to 1000 epochs with a batch size of 73 and 180 iterations per epoch with three fold cross validation. An early stopping and a learning rate adjustment callback technique were used to prevent overfitting and early stagnation. The patience was set to 32 and 5 respectively. The tree models terminated their training after epoch 58, 64, and 63 respectively by the early stopping callback.

A class-weighted categorical focal loss was employed to handle the large imbalance of available samples for the two classes \cite{lin_focal_2018}. The class weights were computed based on the class distributions of the corresponding training subsets of the cross-validation splits.

The three created models were evaluated on the test dataset as an ensemble. The final score was generated by calculating the mean of the softmax output vectors. Afterwards a 0.5 threshold was applied to determine the class. As the primary metric for evaluation balanced accuracy was used as determined by the MIDOG25 challenge for adequate performance comparison of class imbalanced datasets.

The RBR uses a custom feature engineering approach. Instead of providing a classification itself, it returns a modifier, increasing the AMF score and decreasing the NMF score or vice versa. In contrast to the DL approach, the Macenko algorithm \cite{macenko_method_2009} was used for stain normalization. Due to its less aggressive normalization of the blue channel, cells are displayed clearer, proving advantageous in the downstream process. After normalization the pretrained StarDist model "2D versatile he" was used for cell detection \cite{weigert_nuclei_2022}. Cells detected near the image center were selected. In case multiple cells were close to the center and in close proximity to each other, all such cells were included for further analysis.
\begin{itemize}
    \item No cell detected: slight increase in the probability for AMF, as "AMF other" (AMi-Br defined subtype) was often not detected as a cell by StarDist.
    \item One cell detected: analysis of the morphology of the cell. In case of a ring-like, round or oval shape the chance for NMF was increased. Other forms did not adjust the score.
    \item Two cells detected: the orientation of both cells was calculated. If the cells were parallel (difference less than 10°) the prediction was set to NMF. If the cells were close to parallel (difference less than 20°) the change for NMF was significantly increased.
    \item Other cases: no prediction was possible and no adjustment was made
\end{itemize}

Each modification was weighted individually, e.g, a clear parallel orientation increased the chance for NMF by 0.6 (effectively classifying it as NMF regardless of the DL analysis) while a ring-like shape increased the chance for NMF only by 0.2. This does not necessarily represents the significance of the morphology, but rather the trust in the individual analysis algorithms. Each analysis also reported a confidence in its result. Depending on the quality of the analysis and used fallbacks it was possible for certain analyses to be less certain (e.g., determining the orientation of two close to round shapes). If an analysis was not conducted with high confidence, the score was adjusted less significantly.

\begin{figure}[tbp]
    \centerline{\includegraphics[width=0.35\textwidth]{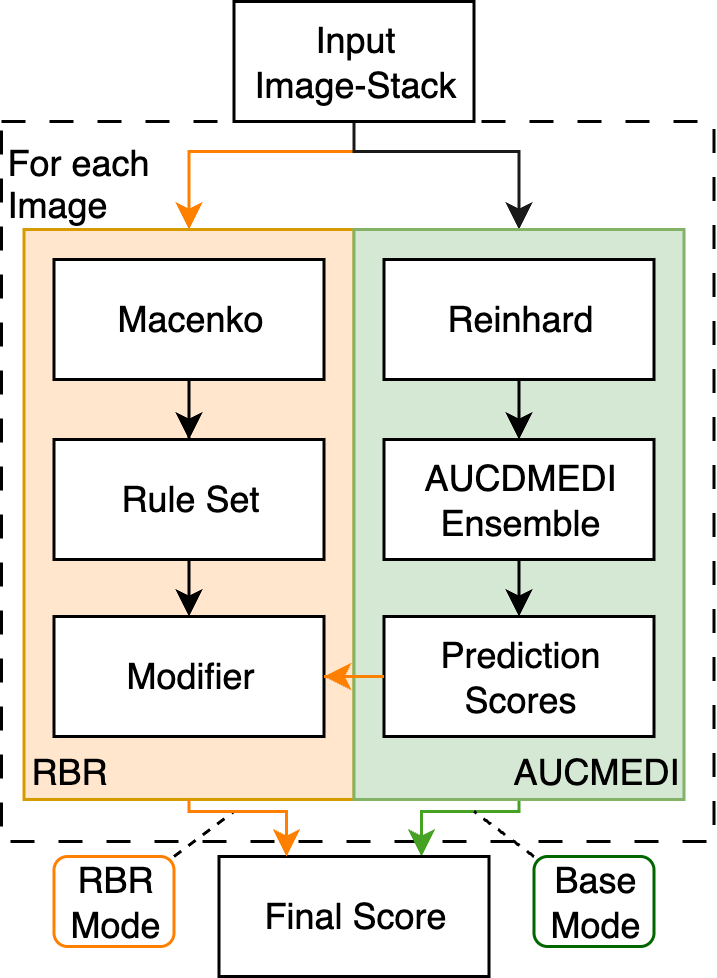}}
    \caption{Inference pipeline of the proposed classification algorithm. If only the AUCMEDI based approach is used the green path is executed, and the AUCMEDI scores are presented as the final result. If the RBR approach is selected, the image-stack is additionally independently analyzed by the rule set and the AUCMEDI score is adjusted based on the analysis.}
    \label{fig:pipeline_deploy}
\end{figure}

Figure \ref{fig:pipeline_deploy} depicts the inference flow. If only the AUCMEDI model is used, each image of the stack is normalized with the Reinhard method, classified by the ensemble, and the ensemble score is reported as the final score. If the RBR mode is selected, the initial process is the same, however, after AUCMEDI finishes its predictions the scores are given to the RBR module. The module normalizes each image with the Macenko algorithm and, when possible, generates a modifier that is subsequently applied to the predicted AUCMEDI score.

For the final submission, a new ensemble was trained on all available data without retaining a local test-split, using the preliminary test set provided in scope of the MIDOG challenge on GrandChallenge for evaluation.

The code for analysis and a documentation to train a new model or adjust the RBR approach is available at GitHub: \href{https://github.com/krausara/MIDOG2025}{https://github.com/krausara/MIDOG2025}

%##############################################################################
\section{Results}
Table \ref{tab:results} depicts the results of the final model on the preliminary test data. The \textit{baseline} model consists of the three ConvNeXtBase ensemble. \textit{RBR} marks the model with the RBR pipeline enabled.

\begin{table}[!ht]
    \caption{Results of the deep ensemble models with and without RBR on the preliminary test set.}
    \label{tab:results}
    \centering
    \begin{tabular}{l|c|c}
          & Baseline & Baseline + RBR\\
        \hline
       Balanced Accuracy & 84.02 & 83.44 \\
       Sensitivity & 92.96 & 85.92 \\
       Specificity & 75.09 & 80.97 \\
       AUC & 92.84 & 89.17 \\
    \end{tabular}
\end{table}

While the use of the RBR module resulted in a worse balanced accuracy, it was able to increase the specificity of the models (Baseline specificity 75.09\% vs. with RBR specificity 80.97\%). However, the increase came at the cost of decreased sensitivity (no RBR 92.96\% vs. RBR 85.92\%), resulting in an overall inferior model.

%##############################################################################
\section{Discussion}
The results confirm the potential of DL ensemble methods for MF classification. The ensemble reached high sensitivity for AMFs, which is crucial for decision support. However, specificity was comparatively low, suggesting that normal figures are more frequently misclassified.

The RBR module demonstrated that domain knowledge can, in principle, improve specificity by incorporating morphological cues such as cell shape and orientation, yet the results were not sufficient for an improvement. An in-depth analysis revealed significant inaccuracies in the MF detection of the general-purpose StarDist model, introduced substantial errors. Further, the manual feature engineering was conducted by computer scientists without pathologists, increasing the likelihood of sub-optimal feature selection.

A key insight is that refinement approaches are not inherently flawed but depend strongly on appropriate upstream models. Dedicated MF-specific detectors could make morphological refinement viable, enabling specificity gains without undermining sensitivity. Furthermore, combining learned and rule-based features remains a promising direction, as it allows balancing data-driven accuracy with interpretable heuristics.

%##############################################################################
\section{Conclusion}
Deep ensembles provide strong baselines for distinguishing atypical from normal mitoses. Rule-based refinement can increase specificity but currently suffers from unreliable cell detection. Future work should focus on MF-specific detectors to enable refinement methods that improve specificity without sacrificing sensitivity.

%##############################################################################
\bibliographystyle{IEEEtran}
\bibliography{IEEEabrv,MIDOG25}

\end{document}